\documentclass[3p,times,twocolumn,nopreprintline]{elsarticle}

\usepackage{amsmath,amsfonts}
\usepackage{algorithmic}
\usepackage{algorithm}
\usepackage{array}
\usepackage[caption=false,font=normalsize,labelfont=sf,textfont=sf]{subfig}
\usepackage{textcomp}
\usepackage{stfloats}
\usepackage{url}
\usepackage{verbatim}
\usepackage{graphicx}

\usepackage{multirow}

\begin{document}

\begin{frontmatter}

\title{Designing Practical Models for Isolated Word Visual Speech Recognition}

\author[1]{Iason Ioannis Panagos\corref{cor1}}
\cortext[cor1]{Corresponding author: I.I. Panagos}
\ead{i.panagos@uoi.gr}

\author[2]{Giorgos Sfikas}
\ead{gsfikas@uniwa.gr}

\author[1]{Christophoros Nikou}
\ead{cnikou@uoi.gr}

\affiliation[1]{organization={Department of Computer Science \& Engineering, University of Ioannina},
	city={Ioannina},
	postcode={45110},
	country={Greece}}

\affiliation[2]{organization={Department of Surveying \& Geoinformatics Engineering, University of West Attica},
	city={Athens},
	postcode={12243},
	country={Greece}}

%\affil*[1]{\orgdiv{Department of Computer Science \& Engineering}, \orgname{University of Ioannina}, \orgaddress{\city{Ioannina}, \country{Greece}}}

%\affil[2]{\orgdiv{Department of Surveying \& Geoinformatics Engineering}, \orgname{University of West Attica}, \orgaddress{\city{Athens}, \country{Greece}}}

%\maketitle

\begin{abstract}
Visual speech recognition (VSR) systems decode spoken words from an input sequence using only the video data.
Practical applications of such systems include medical assistance as well as human-machine interactions.
A VSR system is typically employed in a complementary role in cases where the audio is corrupt or not available.
In order to accurately predict the spoken words, these architectures often rely on deep neural networks in order to extract meaningful representations from the input sequence.
While deep architectures achieve impressive recognition performance, relying on such models incurs significant computation costs which translates into increased resource demands in terms of hardware requirements and results in limited applicability in real-world scenarios where resources might be constrained.
This factor prevents wider adoption and deployment of speech recognition systems in more practical applications.
In this work, we aim to alleviate this issue by developing architectures for VSR that have low hardware costs.
Following the standard two-network design paradigm, where one network handles visual feature extraction and another one utilizes the extracted features to classify the entire sequence, we develop lightweight end-to-end architectures by first benchmarking efficient models from the image classification literature, and then adopting lightweight block designs in a temporal convolution network backbone.
We create several unified models with low resource requirements but strong recognition performance.
Experiments on the largest public database for English words demonstrate the effectiveness and practicality of our developed models.
Code and trained models will be made publicly available.
\end{abstract}


\end{frontmatter}

\section{Introduction}\label{sec:introduction}

Speech Recognition (SR) is the computer vision task which aims to decode speech patterns in digital media.
Sub-tasks of SR are categorized according to the type of the input sequence, more specifically, Audio Speech Recognition (ASR) in the case of audio-only sources, and Visual Speech Recognition (VSR), when the audio is absent.
A system built for automatic SR can utilize any form of recognition depending on the input, for instance, ASR for recordings (no visual component), or VSR for videos without audio, or even combine both forms in a complementary fashion depending on the task, as each type of recognition has its separate benefits and drawbacks.

The rapid adoption of technology in all aspects of everyday life has enabled applications of SR in several domains covering diverse scenarios and requirements.
Notably, SR systems can be utilized in the medical domain, providing support for individuals suffering from an inability to communicate effectively (e.g., speech impairments). 
The widespread usage of mobile devices and computers by a billions of users worldwide has led to the development of accessibility platforms with SR capabilities that offer assistance in times of need and improve the day-to-day lives of people by making their activities and interactions with devices easier and faster.

In entertainment, SR systems have also seen widespread adoption by various video hosting platforms for automatic generation of video transcriptions.
Another application in this industry is found in the archiving and digitization of older films, where SR methods have been employed to generate subtitles without relying on a human expert.
More recently, software offering all forms of SR (audio and/or visual) has been used extensively to assist digital content creators and other individuals, streamlining the process of adding captions to videos, increasing outreach.

These applications of SR mostly fall under the category of ASR, since using only the audio signal for recognition is a comparatively easier task than relying on the visual stream.
Indeed, the spatial dimensions of a video sequence add another layer of complexity in the problem which translates into higher demands in model size and computation overhead that are required to successfully process the more demanding input.
Furthermore, the availability of freely-accessible training data mostly includes corpora originating from audio sources which are easier to collect and transcribe by human listeners, making the process of developing a VSR-only system a more troublesome task due to the volume of data and processing time required.

These factors have contributed in making ASR the predominant type of SR research, with most published methods and software favoring this task over VSR.
For the latter, its role is mostly a complementary one, with usage in special instances where ASR is rendered ineffective either by background noise or audio corruption.
However, due to the additional challenges associated with developing capable VSR systems, this task has remained under active research, with several works focusing exclusively on VSR being published in the last decade.

The spatio-temporal nature of video data (high dimensionality) demands sizable networks to perform accurate recognition of spoken words in video sequences.
The added complexity of this type of input also causes significant data requirements since large-scale datasets are needed to train the models to achieve a high level of performance.
Consequently, works in the recent VSR literature follow the trend of increasing network capacity by proposing larger and more complex models in order to keep improving recognition accuracy.
A drawback of such large architectures is the associated hardware overhead which limits their applicability in practical scenarios by reducing their deployment potential.

The goal of this work is to alleviate this limitation by designing lightweight architectures that still perform competitively with other, more cumbersome ones while being much less demanding on system resources.
To that end, we develop end-to-end models for isolated word VSR that are practical in terms of model size and computation complexity measured in parameter counts and Floating Point OPerations (FLOPs) respectively.
We explore practical, low-cost networks for feature extraction and sequence modeling by analyzing their hardware overhead and recognition performance, adopting the most effective components in an end-to-end architecture.
Our proposed models showcase low hardware demands while achieving high word recognition accuracy, and can be deployed in a wide range of devices to cover more applications and use cases in real-life scenarios.
Extensive experimentation on the largest publicly available corpus for word-level VSR without using any audio data showcases the effectiveness of our compact models in speech recognition of isolated words.

Our contributions can be summarized as:
\begin{itemize}
	\item We explore several lightweight convolutional neural networks from the image classification literature as feature extractors, and benchmark their performance when used in a VSR architecture for word recognition.
	Since this component is responsible for a significant amount of computation of the overall end-to-end architecture, by selecting a robust yet compact model we can achieve savings in model size and computational complexity, without severely compromising performance.
	\item We apply the same method to the sequential model that is used to further process the extracted features, as this component is crucial for strong VSR performance.
	By adapting various network building blocks to an equivalent 1-dimensional causal design, we replace the standard block within a vanilla temporal convolution network to improve its sequence modeling capabilities while keeping size and complexity at affordable levels.
	\item With insights gathered from the above, we design lightweight yet powerful unified architectures and validate their capabilities by performing several experiments and ablation studies on the largest publicly-available dataset for recognition of English words without using any additional training data.
\end{itemize}

We structure the rest of the article as follows:
Section~\ref{sec:related} provides an overview of the contemporary literature on visual speech recognition for isolated words, with an emphasis on approaches using temporal convolution networks and on works that focus on developing practical models.
Our methodology is shown in detail in Section~\ref{sec:proposed}, which begins with a high-level overview of the overall architecture and then expands on each component, showcasing our approach in selecting and evaluating lightweight components.
Section~\ref{sec:experiments} follows, containing an extensive array of experiments and benchmarks in single word recognition alongside model complexity measurements and discussions of results.
Finally, we conclude with Section~\ref{sec:conclusion} offering a summary of the article and discuss insights and directions for future research.

\section{Related Work}\label{sec:related}

The general paradigm that is followed by virtually all works in the published visual speech recognition (VSR) literature decomposes the problem into a series of computation steps.
The first step involves the extraction of intermediate representations as features from the input, while the second models the temporal aspect of the sequence by utilizing these features to further process the relations between them over the duration of the sequence.
Finally, a classifier predicting the output word is added at the end of the network, completing the architecture.

In the last decade, SR architectures typically employ deep Convolutional Neural Networks (CNN) for spatial feature extraction of as well as sequence-to-sequence models for processing the extracted features in a sequential manner.
For the latter task, recurrent architectures based on Gated Recurrent Units (GRU) \citep{liu2020lip, liu2021audio, haq2022using, ai2023cross} and Long Short-Term Networks (LSTM) \citep{stafylakis2017deep, ivanko2022end, xing2023application} have been a popular choice with strong results.

\subsection{TCN-based approaches}\label{sub:tcn_based}

More recently, \textit{Temporal Convolution Networks} (TCN)s \citep{lea2017temporal} have emerged as a viable network suitable for sequence modeling.
At their core, these networks utilize convolutions rather than recurrent cells, and offer higher performance and stable training while not being overly complex.
Due to these attributes, models based on TCNs are being increasingly employed in VSR for sequence modeling, and also in other domains involving data of a sequential or temporal nature, such as in time-series classification.

For isolated word VSR, \citep{martinez2020lipreading} proposed a network based on a TCN that extends the standard architecture with a series of parallel convolutions with multiple temporal scales by using different kernel sizes, increasing the effective receptive field of each block, allowing for greater temporal information flow and improving performance over the base architecture.
This TCN-based model is adopted by \citep{chen2021automatic} and combined with a feature extractor that utilizes hierarchical connections and pyramidal convolution kernels, processing the input at multiple spatial resolutions, while the approach of \citep{sheng2021adaptive} combines the TCN model with a graph convolution network that models the dynamic mouth contours capturing motion and semantic information.
\citep{zhang2022boosting} also uses this multi-scale TCN network of in an architecture that combines a spatial and spatio-temporal view of the input and incorporates information regarding the appearance and shape of the lip region.

A more recent redesign of the temporal convolution block \citep{ma2021lip} borrows the multi-scale structure of \citep{martinez2020lipreading} and adds channel attention and dense connections where all layers within a block are connected.
This method has the advantage of offering rich representations thanks to the dense connectivity, improving performance at the cost of a larger and more complex network in terms of resources.
Due to its success, it has been employed by works such as \citep{chen2023tcs}, where the authors have augmented the densely-connected architecture with several spatio-temporal attention mechanisms for additional feature refinement, further improving accuracy, or adopted as a part of a larger framework for VSR \citep{hao2025lipgen}.

The work of \citep{li2023lipreading} also builds upon the densely-connected network of \citep{ma2021lip} by further refining its block design.
Keeping the multi-scale setup of the base architecture, the model makes use of two branches with multiple convolution sizes each, one extracting local features while the other is more focused on long-range dependencies.
A fusion mechanism to efficiently combine the extracted features is introduced and the output consists of useful representations from each branch connected using a residual pathway.

The authors of \citep{jiang2024gslip} propose a TCN variant that handles local temporal dependencies.
Starting from the architecture of \citep{martinez2020lipreading}, they adopt the same block design where multiple convolutions are used per temporal block, each having a different kernel size.
In contrast to the base design, two branches are used for each kernel size, where one branch applies a regular convolution, and the other a dilated one, following a dilation pattern that increases in each block, gradually enlarging the receptive field to capture more information from the sequence.

\subsection{Lightweight word VSR}\label{sub:lightweight}

Notably, most of the VSR research focus has been directed on improving accuracy leading to developing models with high computational demands which renders their practicality difficult and greatly hinders any potential application in a real-world scenario where resources are limited.
As a result, very few works that consider a practical approach (e.g., developing lightweight models or reducing the size of existing ones) for world-level VSR have been published.

One approach towards this research direction introduced a lightweight end-to-end model for word-level VSR \citep{wisesa2023developing}.
Utilizing a low-cost convolutional neural network for visual feature extraction, the authors combine it with two different sequence modeling networks.
First, a bi-directional GRU is used, and then the authors replace it with a Transformer encoder, further reducing the parameters, the overall complexity and the latency of the model.
In fact, the Transformer encoder even provides a small improvement to the overall accuracy, compared to the recurrent model.

A different approach \citep{panagos2024visual} aims at reducing model sizes by replacing network components in the architecture with lightweight alternatives.
By utilizing shared-parameter drop-in layers in both the feature extraction and the sequence modeling networks, significant reductions in overall model sizes are achieved without greatly affecting performance.
Hyper-parameter tuning can leads to even greater parameter savings with a corresponding drop in accuracy as the model size shrinks.

More recently, \citep{panagos2025lightweight} developed lightweight networks using low-cost network modules that greatly reduce parameter and model complexity measurements.
Practical feature extraction as well as sequence modeling networks are proposed, allowing for a multitude of applications depending on the available resources.
In addition, efficient formulations for the main component of the TCN, the temporal block, are proposed, reducing complexity at the cost of lower accuracy due to the smaller network capacity.

\section{Proposed Model}\label{sec:proposed}

As mentioned in the previous Section, the prevalent approach to tackle the problem of visual speech recognition (VSR) involves a sequence of sub-tasks.
Rather than attempting to solve the entire problem at once, a separate module is employed to handle each sub-task.
This paradigm offers a higher degree of design flexibility, since each sub-task has a different objective and a specialized architecture that performs best for the particular sub-task may be chosen.

This design can be summarized as:
\begin{align}
	\label{eq:vsr_system}
	f = Feature\_extraction(i) \nonumber \\
	s = Sequence\_modeling(f) \nonumber \\
	o = Classification(s),
\end{align}
\noindent where $i$ represents the input sequence, and $o$ is the output word of the network.

The high-level overview of a VSR system in Eq.~\ref{eq:vsr_system} outlines three distinct steps when processing an input sequence.
The first two are often handled by neural network models as they offer strong performance for these two tasks.
Simultaneously, these are the most computationally intense modules in the entire architecture.
When designing a lightweight VSR system, component selection plays a crucial role as their size and structure (i.e., amount of blocks and operations used, including other hyper-parameters) determine the computational overhead of the overall architecture.

\begin{figure*}[ht]
	\centering
	\includegraphics[width=\linewidth]{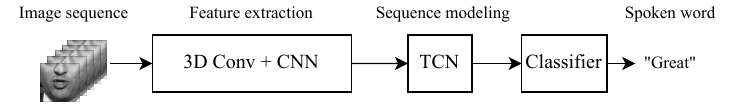}
	\caption{Overview of the architecture used for visual speech recognition. We experiment with several feature extractors as well as our proposed lightweight sequence models based on a TCN backbone. For final assignment, the Softmax function is used. The system outputs a spoken word.}
	\label{fig:model}
\end{figure*}

Figure~\ref{fig:model} illustrates the general design described in Eq.~\ref{eq:vsr_system}.
In the following Subsections, we construct a multitude of architectures utilizing a variety of lightweight networks for the task of feature extraction and various efficient block designs for the task of sequence modeling.
We benchmark these architectures on the LRW test set (Section~\ref{sec:experiments}), finding which architectures perform best on this data set while being lightweight in terms of resources (FLOPs and network parameters).

\subsection{Visual Feature Extraction}\label{sub:vfe}

The visual feature extraction step involves spatial processing of the input sequence, generating a set of intermediate representations with a high channel dimensionality.
The convolutional networks employed in this step are designed in stages, where the operations in each stage gradually reduce the spatial size of the input while simultaneously increasing the channel dimension.

In this work, our focus is developing lightweight end-to-end architectures for isolated word VSR, therefore a highly efficient feature extractor is an integral component of such an architecture.
With that in mind, we experiment with several lightweight models proposed in the image classification literature, since we find that these networks tend to perform well as feature extractors.
This selection of networks covers a diverse range of approaches in terms of network and block design and can showcase which lightweight model approach is more suited to the task of feature extraction in the context of visual speech recognition.
More specifically, we experiment with the following networks:

\begin{itemize}
	\item \textit{MobileNetV2} \citep{sandler2018mobilenetv2} was selected for its high performance and low computation cost.
	It utilizes an \textit{inverted residual} block design that shifts the connectivity of shortcuts and utilizes a point-wise convolution to increase the amount of channels, as opposed to a standard residual block.
	\item \textit{MobileNetV4-S} \citep{qin2024mobilenetv4} is a recently-proposed model resulting from a Neural Architecture Search (NAS) process, meaning that its structure was searched automatically using an algorithm rather than being designed by human researchers.
	MobileNetV4 represents the latest iteration of this family of networks, incorporating a series of innovations regarding block design and was selected to explore the performance and adaptability of an architecture for image classification that was searched with an objective that balances the trade-off between latency and performance.
	\item \textit{EMO-1M} \citep{zhang2023rethinking} introduces a building block based on the inverted residual design of MobileNetV2 combined with self-attention.
	\item Next, we also choose \textit{InceptionNeXt-A} \citep{yu2024inceptionnext}, which combines modern block design principles \citep{liu2022convnet} with three parallel depth-wise convolutions inspired by the Inception \citep{szegedy2016rethinking} block.
	To keep computation overhead manageable, the block splits its input to equal parts applies each convolution to a different chunk, retaining a skip connection and concatenates the outputs.
	\item Finally, \textit{StarNet-050} \citep{ma2024rewrite} presents an alternative approach to efficient network design by adopting the (element-wise) multiplication operator to combine high dimensional features within a building block.
\end{itemize}

Each CNN is superseded by a small stack of 3D convolution, Batch Normalization and non-linear activation layers, which serves as a spatio-temporal processing unit extracting more short-term dependencies from the input.
This convolution uses a 3-D kernel shape of $(3, 5, 5)$, where $3$ corresponds to the temporal dimension and $5, 5$ to the spatial, with an output of $32$ channels and its computational overhead is marginal compared to the other components of the overall pipeline.
Using this small stack is common practice for the task of VSR, e.g., \citep{stafylakis2017deep, martinez2020lipreading, ma2021lip, panagos2024visual, panagos2025lightweight}, where an additional pooling layer is added to further reduce the spatial dimensions, also keeping computations low.
We find that when using lightweight networks, the pooling layer does lower the network size and computational complexity, but significantly harms recognition performance and for this reason pooling is not used in our models.

\subsection{Sequence Modeling Network}\label{sub:smn}

The sequence modeling network ingests the features produced by the previous step for further processing by modeling the temporal aspect of the input sequence.
The goal is discovering the inter-relationships that exist between features across the length of the sequence, since during complex speech (words, phrases and sentences) the movements of the mouth follow sequential patterns of motion.
Capturing this information can lead to higher accuracy when making a prediction.

Related research in the VSR literature has demonstrated that improving the sequence modeling network can bring significant benefits in word recognition performance, as seen by the improvements obtained by the works of \citep{martinez2020lipreading} and \citep{ma2021lip}, where the same visual feature extractor network is used.
The higher performance obtained by these two works can be attributed to replacing recurrent networks with TCN-based models and then modifying the temporal convolution blocks with more powerful designs.
In a similar fashion, we aim to develop a lightweight but powerful sequence modeling network that keeps the overall computation at affordable levels while raising accuracy.
For this reason, we adopt the TCN formulation as the backbone structure for sequence modeling and explore several block designs borrowed from lightweight CNNs.
To adapt these blocks to our task of VSR, we convert all 2-D layers (i.e., convolution and normalization) to 1-D.

More specifically, we use:
\begin{itemize}
	\item The \textit{Linear} layer \citep{xu2022etinynet} which utilizes two depth-wise convolutions with a point-wise in-between which is used to fuse the information from the channels,
	\item the \textit{Fused MB} \citep{tan2021efficientnetv2} that relies on a regular convolution to expand the channel dimensionality and a point-wise convolution to mix the channels,
	\item the \textit{Inverted Residual} \citep{sandler2018mobilenetv2} layer, that reverses the order of operations of the Linear layer (i.e., uses a depth-wise convolution between two point-wise ones) and has been a popular building block in several lightweight architectures, e.g., \citep{tan2019mnasnet, tan2019efficientnet} as well as a starting point for other efficient blocks, e.g. \citep{howard2019searching, liu2022convnet, wang2024yolov10, qin2024mobilenetv4}.
	\item The recently-proposed \textit{UIB}, introduced in \citep{qin2024mobilenetv4}, which is an advanced and flexible efficient block that is employed on the MobileNetV4 family of models \citep{qin2024mobilenetv4}.
	\item The also recently-proposed \textit{CIB} \citep{wang2024yolov10}, representing advances in convolution-based lightweight blocks for compact networks and is the basic building block of the YOLOv10 backbone \citep{wang2024yolov10},
	\item And finally, the \textit{Star} block (variant \textit{V}), proposed in \citep{ma2024rewrite} that exploits the multiplication operation to explore how this approach performs for the task of VSR.
\end{itemize}

\begin{figure*}[!ht]
	\centering
	\includegraphics[width=.8\linewidth]{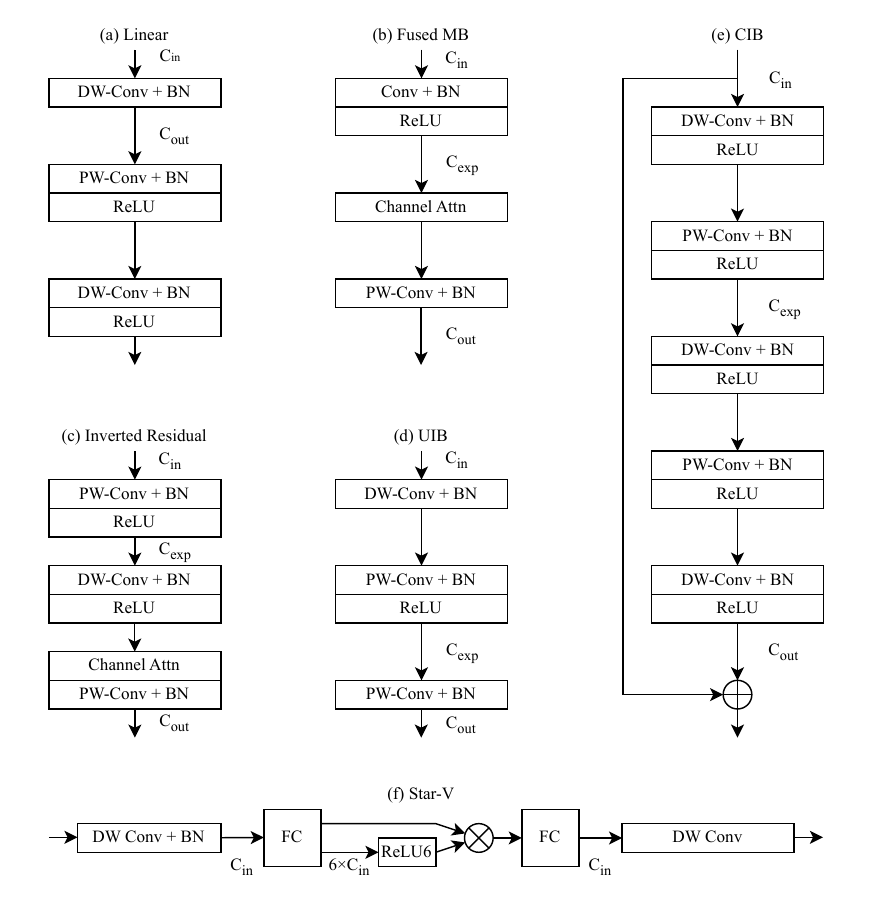}
	\caption{Structural illustration of the lightweight building blocks that were used in this work. Convolution and Batch Normalization layers were converted to 1-D. $C_{in}$, $C_{exp}$ and $C_{out}$ refer to input, expanded and output channels within a block. $\bigotimes$ denotes element-wise multiplication (Hadamard product).}
	\label{fig:blocks}
\end{figure*}

Diagrams showing the layers of the blocks are shown in Figure~\ref{fig:blocks}.

\section{Experiments}\label{sec:experiments}

\subsection{Dataset \& Preprocessing}\label{sub:dataset}

The models proposed in this work are trained and evaluated on the LRW\footnote{\url{https://www.robots.ox.ac.uk/~vgg/data/lip_reading/lrw1.html}} \citep{chung2017lip} dataset, which is currently the largest openly-available corpus for visual speech recognition of isolated words in-the-wild.
It is comprised of short clips where one word is uttered at the middle of the frame sequence, with a vocabulary of $500$ distinct words.
The clips are recorded from public television programs such as news shows and feature significant speaker and background variety with adequate illumination.
Moreover, frontal and sideways views of the speakers are also included, increasing the complexity of the dataset.

The corpus is split in a train, a validation and a test sub-set, and frame sequences in each set do not overlap.
The train set contains $488.766$ clips, while the validation and test set contain $25.000$ samples each.
The length of each clip is fixed at $29$ frames in a spatial resolution of $256\times256$.
In total, the summed length of all clips in all sub-sets of the LRW dataset is $173$ hours.

For pre-processing, a series of steps are performed on the raw data. 
After detecting the speaker's face in each frame, a face alignment network is employed to compute RoI landmarks.
Then, normalization of the images occurs by removing variations of size and rotations using a mean face shape.
A bounding box of shape $96\times96$ crops the region around the speaker's mouth area, which is further normalized by mean and standard deviation and finally converted to gray scale to remove color, resulting in a simpler (from a computational standpoint) final image.
This procedure is typically employed by works in the literature using LRW (e.g., \citep{martinez2020lipreading, ma2021lip}).

\subsection{Training Setup}\label{sub:training}

We use PyTorch to develop our models and perform all experiments in this Section.
All network weights are initialized randomly without using any pre-trained checkpoints, and all models proposed in this work use the following configuration.
The LRW training set is used, and after each epoch the model is validated on the validation set.
We train for a fixed amount of $80$ epochs in total, saving the best-performing (in the validation set) checkpoint.
At the end of training, the best-performing weights are loaded and the model is evaluated on the LRW test set.
Stochastic Gradient Descent with an initial learning rate of $0.02$ is used to update the weights.
A decay factor of $0.01$ is applied to all weights to prevent over-fitting.
Further regularization in the form of a fixed dropout rate of $0.2$ is also added to the TCN layers, as it was found to improve performance over not using any amount of dropout.

The GPUs used to train and evaluate the models are Nvidia RTX 2080Ti with $11$ GB of VRAM, therefore we use a batch size of $32$ which allows fitting an entire model to a single GPU.
The learning rate scheduling followed is cosine annealing, which has been found to perform very well for this dataset:
\begin{equation}
	\label{eq:cosine}
	\eta_t = \eta_{min} + \frac{1}{2} (\eta_{max} - \eta_{min})\left(1 + \cos\left(\pi\frac{T_{cur}}{T_{max}}\right)\right),
\end{equation}
\noindent where $\eta_t$, $\eta_{min}$, $\eta_{max}$ are the current, minimum and maximum values of the learning rate, respectively, while $T_{cur}$ and $T_{max}$ measure the number of iterations.
In our case, we simplify the above Equation by setting the amount of epochs as the iteration number, $0$ as the minimum rate and $1$ as the maximum, respectively, therefore Eq.~\ref{eq:cosine} becomes:
\begin{equation}
	\label{eq:cosine_simple}
	\eta_t = \frac{1}{2}\left(1 + \cos\left(\pi\frac{epoch}{80}\right)\right),
\end{equation}
with $epoch$ referring to the current epoch in the training process (and the value $80$ corresponds to the total epochs).
This annealing rate is applied after every epoch.

During training, the following augmentations are employed:
\begin{itemize}
\item random spatial crop and flip of the frame,
\item MixUp \citep{zhang2018mixup} with an $\alpha$ of $0.4$, and
\item variable length augmentation \citep{martinez2020lipreading}, following previous works (e.g., \citep{ma2021lip, panagos2024visual}).
\end{itemize}

\section{Results and Discussions}\label{sec:results}

In keeping up with the structure of Section~\ref{sec:proposed}, the results and discussions presented in this Section follow the same order, starting with the feature extraction experiments and followed by an evaluation of the temporal convolution blocks.
To measure parameter and Floating Point OPerations (FLOP)s, the \textit{torchinfo}\footnote{\url{https://github.com/TylerYep/torchinfo}} python package is used.
All measurements are obtained using a single sequence as the input ($29$ frames with $88\times88$ resolution), to simulate applying the architecture to a video of the LRW test set in an in-the-wild scenario, giving us an accurate representation of real-world resource requirements.
In addition, all models are trained using the same setup as described in Section~\ref{sub:training}, to keep comparisons fair.

\subsection{Feature Extraction}\label{sub:fe_results}

To benchmark the lightweight feature extractors, we employ a simple TCN-based sequential model following the architecture of \citep{bai2018empirical}.
This model uses $4$ main stages, each containing a block with a sequence of temporal convolution, batch normalization and rectified linear unit as an activation functions, repeated twice.
The kernel size for all convolutions is set to $3$ while the input and output channels are set to $512$.
The dilation rate used by both convolutions in each block is exponentially increased at every stage (to $2^{stage}$), starting from $1$.
To keep comparisons fair and meaningful, we use this TCN sequence model in combination with each lightweight feature extraction network discussed earlier, creating different end-to-end architectures.
As a baseline for comparison, we use an 18-layer residual network \citep{he2016deep} (also using this TCN configuration), which is favored by many VSR works for its strong performance (e.g., \citep{stafylakis2017deep, martinez2020lipreading, ma2021lip, panagos2024visual}) at the cost of very high complexity (compared to the lightweight networks).
Results are presented in Table~\ref{tab:fe_results}.

\begin{table}[!th]
	\caption{Benchmark results of lightweight feature extractors on the LRW test set compared against the much larger and more expensive residual network baseline. Each model is combined with a 4-layer TCN.}
	\label{tab:fe_results}
	\centering
	\resizebox{\linewidth}{!}{
	\renewcommand{\arraystretch}{1.1}
	\begin{tabular}{l c c c}
		\hline
		\multirow{2}{*}{\textbf{Model}} & \textbf{FLOPs ($\times 10^9$)} & \textbf{Params ($\times 10^6$)} & \textbf{Accuracy} \\
		& \textbf{Total (CNN)} & \textbf{Total (CNN)} &  \textbf{$\uparrow$ (\%)}\\
		\hline
		ResNet baseline (18-layer) \citep{he2016deep}   & 31.2 (30.8) & 17.7 (11.1) & 87.7 \\
		\hline
		MobileNetV2 \citep{sandler2018mobilenetv2}      &  0.9 (0.5) &  8.5 (1.9) & 82.9 \\
		MobileNetV4-S \citep{qin2024mobilenetv4}        &  1.9 (1.5) &  7.7 (1.2) & 84.8 \\
		InceptionNext-A \citep{yu2024inceptionnext}     &  0.8 (0.4) &  9.6 (3.0) & 83.1 \\
		EMO-1M \citep{zhang2023rethinking}              &  1.4 (1.1) &  7.8 (1.2) & 83.6 \\
		StarNet-050 \citep{ma2024rewrite}               &  0.7 (0.4) &  7.0 (0.4) & 82.7 \\
		\hline
	\end{tabular}
	}
\end{table}

Unexpectedly, the lightweight feature extractors achieve lower complexity and model size at the cost of performance.
The MobileNetV4-S \citep{qin2024mobilenetv4} model is the strongest performer over all the lightweight architectures at $84.8\%$ accuracy, $2.9\%$ lower than the baseline.
While its network size is smaller than other models, e.g., InceptionNext-A \citep{yu2024inceptionnext}, its complexity in terms of FLOPs is the highest after the baseline, which is arguably the reason it achieves a higher accuracy.
A similar behavior can be noted for the EMO-1M \citep{zhang2023rethinking} and StarNet-050 \citep{ma2024rewrite} models, which are close in performance with similar parameter counts and slightly lower complexity (FLOPs). 
The InceptionNext-A model measures higher in parameters compared to the others due to its multi-convolution design which nevertheless keeps the operations rather low, which could be a reason for its performance, similarly to the StarNet-050 model.

All lightweight feature extractors reduce the overall size by an amount ranging from $8.1$ ($45\%$) to $10.7$ ($60\%$) million parameters.
A more remarkable improvement lies in the overall computational complexity of the models, which is explained by the more efficient block designs and amounts to a $94\%$ reduction in the case of MobileNetV4-S (the highest FLOP count and performance among the lightweight models) and $98\%$ for StarNet-050 (which achieves the lowest results in both metrics).
The remaining models in our list still achieve reductions $\ge95\%$ in complexity.
For the purposes of this work, the MobileNetV4-S model achieves the highest accuracy over the other lightweight models, $1.2\%$ higher than EMO (which comes second in performance) at very similar FLOP and parameter counts, and will be used as the feature extractor in the following experiments.

\subsection{Temporal Convolution Blocks}\label{sub:tc_resuls}

Having benchmarked the lightweight networks for visual feature extraction, we shift our attention to the sequence modeling component of the architecture.
All experiments in this Subsection use a unified model that combines a MobileNetV4-S, which is the best-performing lightweight feature extractor, with a TCN-based architecture that employs a customized temporal convolution block described in detail in Section~\ref{sub:smn}.
The comparison is shown in Table~\ref{tab:tc_results} and evaluates how the different lightweight block designs perform in the task of VSR of isolated words when combined with a low-resource convolutional feature extractor.
Simultaneously, it showcases which of the blocks can be used to recover some performance that is lost due to the smaller networks, creating an architecture that is compact but performs competitively.

\begin{table}[!th]
	\caption{Benchmark results of using different temporal block configurations with the MobileNetV4-S feature extraction architecture on the LRW test set compared against the baseline TCN \citep{bai2018empirical}.}
	\label{tab:tc_results}
	\centering
	\resizebox{\linewidth}{!}{
	\renewcommand{\arraystretch}{1.1}
	\begin{tabular}{l c c c}
		\hline
		\multirow{2}{*}{\textbf{Temporal Block}} & \textbf{FLOPs ($\times 10^9$)} & \textbf{Params ($\times 10^6$)} & \textbf{Accuracy} \\
		& \textbf{Total (TCN)} & \textbf{Total (TCN)} & \textbf{$\uparrow$ (\%)} \\
		\hline
		TCN block (baseline) \citep{bai2018empirical} & 1.9 (0.2) & 7.7 (6.2) & 84.8 \\
		\hline
		Linear \citep{xu2022etinynet} & 1.7 (0.03) & 2.5 (1.0) & 83.5 \\
		FusedMB \citep{tan2021efficientnetv2} & 2.1 (0.5) & 16.1 (14.6) & 86.8 \\
		Inverted Residual \citep{sandler2018mobilenetv2} & 1.8 (0.1) & 5.6 (4.2) & 83.4 \\
		CIB \citep{wang2024yolov10} & 1.7 (0.1)  & 5.7 (4.2) & 86.7 \\
		UIB \citep{qin2024mobilenetv4} & 1.9 (0.2) & 9.8 (8.4) & 86.8 \\
		Star-V \citep{ma2024rewrite} & 2.0 (0.3) & 14.0 (12.6) & 88.1 \\
		\hline
	\end{tabular}
	}
\end{table}

In this evaluation, we can see that the Linear \citep{xu2022etinynet} and Inverted Residual \citep{sandler2018mobilenetv2} blocks cause a drop of $1.3\% - 1.4\%$ in performance, arguably the result of reducing the network parameters, since the FLOPs are largely unaffected ($0.1 - 0.2$ GFLOP reduction).
Of these two blocks, the Linear is more efficient and suitable for very low resource scenarios, reducing the baseline parameters by more than $3\times$, and achieving a slightly higher accuracy than the Inverted Residual.
The FusedMB \citep{tan2021efficientnetv2} block performs similarly to the CIB \citep{wang2024yolov10} and UIB \citep{qin2024mobilenetv4} blocks with nearly the same accuracy, however it uses a regular convolution which, when combined with a high expansion ratio, raises the overall parameters and complexity of the network (more than $2\times$ that of baseline).
The CIB block reduces both FLOPS and parameters and simultaneously raises the accuracy by $1.8\%$, which can be attributed to its more modern design and can be considered as an upgrade to the standard TCN without any drawbacks, and is also a strong candidate when designing a lightweight solution.
The UIB block performs identically to the FusedMB but is a preferable choice as it is lighter on parameters (by about $39\%$) as well as FLOPS ($0.2$ fewer GFLOPs), and when compared to the standard TCN block it raises the parameter count by $2.1$ M and the accuracy by $2.0\%$, results that can be attributed to its modernized design, similar to CIB.
Nevertheless, the CIB block is a more efficient choice than UIB as it almost matches its performance ($0.1\%$ difference), but with fewer GFLOPs and parameters ($0.2$ and $4.1$M).

In these experiments, the best performing design is the Star-V \citep{ma2024rewrite} block, which surpasses the baseline by $2.7\%$ accuracy at the cost of a larger size ($1.8\times$ more parameters) due to the large kernels used by its convolutions.
Even so, the computational overhead of this model is maintained at manageable levels ($0.1$ GFLOPs more than baseline), since the convolutions used are depth-wise.
The combination of a TCN with Star-V blocks and a MobileNetV4-S feature extractor achieves the highest accuracy in this comparison, $0.4\%$ more than the much larger and computation-heavy baseline (of Table~\ref{tab:fe_results}) while being $10$ million parameters smaller and having an impressive $15.65\times$ fewer FLOP count.

Next, we perform another round of benchmarking of the lightweight feature extractors as well as the high-resource residual baseline from the previous Section by combining them with a TCN using the best-performing temporal block (Star-V) and tabulate the results in Table~\ref{tab:fe_sv_results}.

\begin{table}[!th]
	\caption{Combining the TCN with Star block (variant V) \citep{ma2024rewrite} with other visual feature extraction networks. FLOP and parameter measurements include both components.}
	\label{tab:fe_sv_results}
	\centering
	\resizebox{\linewidth}{!}{
	\renewcommand{\arraystretch}{1.1}
	\begin{tabular}{l c c c}
		\hline
		\textbf{Temporal Block} & \textbf{FLOPs ($\times 10^9$)} & \textbf{Params ($\times 10^6$)} & \textbf{Acc. $\uparrow$ (\%)} \\
		\hline
		ResNet (18-layer) \citep{he2016deep}         & 31.3 & 24.0 & 90.0 \\
		\hline
		MobileNetV2 \citep{sandler2018mobilenetv2}   & 1.0 & 14.8 & 87.0 \\
		InceptionNext-A \citep{yu2024inceptionnext}  & 0.9 & 15.9 & 86.6 \\
		EMO-1M \citep{zhang2023rethinking}           & 1.6 & 14.1 & 86.7 \\
		StarNet-050 \citep{ma2024rewrite}            & 0.9 & 13.3 & 87.0 \\ 
		\hline
	\end{tabular}
	}
\end{table}

Compared to the standard TCN design used to benchmark the feature extractors (Table~\ref{tab:fe_results}), using the Star-V block can raise performance by up to $4.3\%$ while only adding $0.2$ GFLOPs which is the case for StarNet-050 \citep{ma2024rewrite}, making this combination ideal for situations with rather constrained computational capabilities (e.g., edge devices).
We observe a similar outcome for all other lightweight models, where non-trivial raises in accuracy are achieved, demonstrating that using this block can recover lost performance with a minimal impact on computation.
Invariably, the number of parameters is increased due to the design of the Star-V block (see Figure~\ref{fig:blocks}), as explained previously, representing a trade-off with the improvement in network accuracy, which some applications might find an acceptable compromise.
When paired with the larger and more powerful 18-layer residual network it achieves a $2.3\%$ improvement over the standard TCN, reaching a $90.0\%$ recognition accuracy.
These results also demonstrate that the Star-V block scales well with all networks regardless of their size and complexity and establish it as a powerful building component for a sequence modeling network when designing efficient end-to-end models for VSR.

\subsection{Comparison With Other Methods}\label{sub:sota_results}

Finally, we compare our best-performing models with other approaches from the literature on the task of word-level VSR on the LRW dataset.
Since we do not use additional training data, word boundaries or audio cues, for a fair comparison, we compare with models that meet these criteria.
The reasoning behind this choice is that using additional training data such as extra data sets or video sequences is not feasible for several languages taking into account the additional effort required to collect and annotate the data, not to mention the additional training time which can be a factor in some applications.
Similarly, word boundaries indicate the frame where the word is present in the video clip, which is additional information that is not available in-the-wild and thus does not reflect real-world conditions.
As for audio, while some works utilize the audio stream in architectures that leverage both audio and video modalities, in this work we consider it out of scope as we develop a video-only approach since the audio is not always available (e.g., in silent footage).
For each method, we also include size and computational complexity measurements, providing a more complete comparison between the different methods.

\begin{table*}[!th]
	\caption{Comparison of our method (highlighted) with recent works from the word VSR literature. FLOP and parameter measurements include both components.}
	\label{tab:results}
	\centering
	\setlength{\tabcolsep}{12pt}
	\begin{tabular}{l@{}c c c}
		\hline
		\textbf{Model architecture} & \textbf{FLOPs ($\times 10^9$)} & \textbf{Params ($\times 10^6$)} & \textbf{Accuracy $\uparrow$ (\%)} \\
		\hline
		ResNet + MS-TCN \citep{martinez2020lipreading} & 10.31 & 36.4 & 85.3 \\
		ResNet + DC-TCN \citep{ma2021lip} & 10.64 & 52.54 & 88.3 \\
		ResNet (G) + MS-TCN (G) \citep{panagos2025lightweight} & 3.60 & 16.73 & 86.67 \\
		ResNet (G) + DC-TCN (G) \citep{panagos2025lightweight} & 3.85 & 29.48 & 87.58 \\
		ResNet + DC-TCN (G) \citep{panagos2025lightweight} & 10.01 & 37.81 & 89.10 \\
		
		ShuffleNet v2 (1$\times$) + MS-TCN \citep{ma2021towards} & 2.23 & 28.8 & 85.5 \\
		ResNet + 3$\times$Bi-GRU \citep{feng2021efficient} & 10.54 & 59.5 & 88.4 \\
		ResNet + 2$\times$Bi-LSTM \citep{ivanko2022visual} & 10.24 & 50.07 & 88.7 \\

		\hline
		\bfseries MobileNetV4 + TCN (Star-V block) & 2.0 & 14.0 & 88.1 \\
		\hline
	\end{tabular}
\end{table*}

Compared to other methods from the VSR literature, our model achieves slightly lower recognition performance, which can be attributed to its smaller size and complexity.
Since most architectures in this comparison utilize a residual network baseline, our model costs $5\times$ fewer FLOPs, as it uses the much smaller MobileNetV4 feature extraction network, which also saves about $10$ M parameters.
Simultaneously, the TCN with Star-V blocks is much more compact than the recurrent architectures or the larger TCN variants that use multiple convolutions per block.
The model of \citep{ma2021towards} is comparable in overhead to our work but is surpassed in accuracy, while being more than double in overall size, and the same applies to the lightweight networks proposed in \citep{panagos2025lightweight}.
In fact, regarding parameter counts, our end-to-end model is the smallest one in this comparison and also the most efficient in terms of complexity, yet, in spite of its small size, it falls behind some of the larger models by only about $1.0\%$ accuracy.
These results showcase our proposed model's strong performance at minimal size and network complexity, attributes that make it an ideal choice for applications where a highly efficient model is needed.

\subsection{Ablation Studies}\label{sub:ablations}

\textbf{Star blocks.}
The work of \citep{ma2024rewrite} introduces several variants for the \textit{Star} block with similar architectural designs and common characteristics.
All blocks start and end with a depth-wise convolution operation and include point-wise convolutions that expand and subsequently restore the input channels according to a fixed expansion rate.
Another similarity is the use of the RELU6 function for non-linearity and the multiplication operation for feature mixing.
The reader is referred to the Supplementary material of \citep{ma2024rewrite} for details on the architectures of each Star block variant.
In our case, we convert each variant to 1-D for use in the TCN and train separate models, presenting the results in Table~\ref{tab:ablation_s}.
A MobileNetV4-S is used for feature extraction.

\begin{table}[!th]
	\caption{Ablation study on the architecture of the Star block \citep{ma2024rewrite} used in the TCN model. For size and complexity, only the TCN is measured.}
	\label{tab:ablation_s}
	\centering
	\resizebox{\linewidth}{!}{
	\begin{tabular}{l c c c}
		\hline
		\textbf{Block} & \textbf{FLOPs ($\times 10^9$)} & \textbf{Parameters ($\times 10^6$)} & \textbf{Accuracy $\uparrow$ (\%)} \\
		\hline
		Star-I    & 0.36 & 12.6 & 87.9 \\
		Star-II   & 0.36 & 12.6 & 87.0 \\
		Star-III  & 0.73 & 25.2 & 88.1 \\
		Star-IV   & 0.36 & 12.6 & 87.9 \\
		Star-V    & 0.36 & 12.6 & 88.1 \\
		\hline
	\end{tabular}
	}
\end{table}

The best-performing designs of the Star block are variants III and V, achieving the same accuracy, while variants I and IV are following closely.
In this comparison, variant II achieves the lowest performance at $87.0\%$, which is still higher than the other lightweight blocks (see Table~\ref{tab:tc_results}).
In terms of complexity and parameters, since all variants share a similar architecture, we notice identical measurements in FLOPS and parameters with the exception of variant III, which applies an additional point-wise convolution after expanding channels, causing the increase in parameters and FLOPS.
Given these measurements, the best choice for the task of VSR is variant V, which achieves the highest performance while being as efficient as the other variants.

\textbf{Architecture configuration.}
The previous experiments used a temporal convolution network with a four-stage design and $512$ channel outputs, following \citep{ma2021towards}.
This amount of stages is used, to the best of our knowledge, by virtually all works in the VSR literature that employ a TCN model for sequence modeling, since they typically adopt the models of \citep{martinez2020lipreading} or \citep{ma2021towards} that are also four-stage architectures.
Similarly, convolutional networks (e.g., ResNets \citep{he2016deep}, MobileNets \citep{sandler2018mobilenetv2}) also commonly use four stages in their designs. 
Regarding the amount of channels in each block, the use of $512$ is empirical, striking a balance between complexity, size and accuracy.
In Table~\ref{tab:ablation_a} we show the results of an ablation study on the stages and channels of the TCN, experimenting with architectures that are shallower (fewer stages with more channels per stage to compensate) or deeper (more stages with fewer channels, respectively).
As before, a MobileNetV4-S is used for feature extraction.

\begin{table}[!th]
	\caption{Ablation study on the configuration of the TCN architecture. For size and complexity, only the TCN is measured.}
	\label{tab:ablation_a}
	\centering
	\resizebox{\linewidth}{!}{
	\begin{tabular}{c c c c}
		\hline
		\textbf{Configuration} & \textbf{FLOPs} & \textbf{Params} & \textbf{Accuracy} \\
		\textbf{Stages $;$ Channels / stage} & ($\times 10^9$) & ($\times 10^6$)& $\uparrow$ (\%) \\
		\hline
		2 ; 1024 & 0.51 & 17.8 & 86.0 \\
		3 ; 768  & 0.53 & 18.5 & 87.2 \\
		4 ; 512  & 0.36 & 12.6 & 88.1 \\
		6 ; 256  & 0.18 & 6.4 & 87.6 \\
		8 ; 128  & 0.10 & 3.4 & 86.2 \\
		\hline
	\end{tabular}
	}
\end{table}

This ablation study showcases that a network with four stages and $512$ channels is the best approach for the task of VSR as the baseline configuration outperforms all other approaches.
Making the network shallower by reducing the number of stages gradually degrades performance as more stages are required to process the information, while compensating by increasing the channels in the convolutions in each temporal block leads to an increase in network overhead.
Similarly, making the network deeper by adding stages also lowers its performance (albeit by a lower amount than removing stages), which can be attributed to two factors: 
lowering the amount of channels per layer, thus hampering the expressiveness of the network and its ability to capture information, and impeding backward gradient flow during training since the back-propagation path becomes longer.
The deeper networks perform marginally better than the very shallow ones, while being much more efficient in terms of overhead and are suitable for special cases with high resource restrictions.

A key attribute of the TCN design is the dilation factor used in each convolution which increases with every block.
It is therefore possible that in the case of shallow networks (dilation$=\{1,2\}$) this rate is rather low and could be another cause for the lowered performance since the second block does not process a broader amount of information, potentially missing key temporal relationships from neighboring frames.
For the deeper networks, the later stages employ high rates of dilation (since it is doubled at every block) which allows them to cover temporally distant information that might be irrelevant to the current frame.

Investigating these hypotheses is left as future work.

\section{Conclusion}\label{sec:conclusion}
In this work, we presented a systematic approach into designing lightweight architectures for practical visual speech recognition applications.
A VSR system extracts spoken words from the input sequence in a two distinct computation steps, visual feature extraction and temporal sequence modeling, each handled by a different component.
For visual feature extraction, we benchmarked several lightweight models from the image classification literature, finding that significant reductions in FLOPS ($ \ge94\%$) and parameter counts ($\ge45\%$) are achievable, compared to the expensive baseline network, at the cost of lowering recognition accuracy.
For modeling the temporal aspect of the extracted features, a multitude of efficient designs were explored as temporal block replacements in a standard TCN model, taking advantage of its favorable properties and high performance for the task of VSR.
All models were trained and evaluated on the largest public dataset for word VSR in English and our findings show that when an efficient feature extraction model is combined with a robust sequence model, significant gains in accuracy are possible, mitigating the losses that occur from lowering the network capacity.
We also find that the Star block which takes advantage of the multiplication operator is a very strong performer in the task of single word VSR, demonstrating impressive performance at low network overhead, that also scales favorably with larger networks.
In fact, our most efficient model combines a MobileNetV4 with a TCN using Star-V blocks and is very competitive with other much larger methods from the VSR literature.
Future work involves exploring techniques to further improve recognition accuracy, bridging the performance gap with larger models.

%\backmatter

\bibliographystyle{elsarticle-num-names}
\bibliography{bibliography}

\end{document}